\pdfoutput=1
\documentclass[11pt]{article}
\usepackage[]{acl}

\usepackage{times}
\usepackage{verbatim}
\usepackage{latexsym}
\usepackage{graphicx}
\usepackage{booktabs}
\usepackage{multirow}
\usepackage{float}

\usepackage[T1]{fontenc}

\usepackage[utf8]{inputenc}
\usepackage{amsmath}
\usepackage{cleveref}

\usepackage{microtype}
\usepackage{booktabs}
\usepackage{multirow}

\usepackage{inconsolata}
\title{Investigating Mysteries of CoT-Augmented Distillation}

\author{\textbf{Somin Wadhwa}\quad\quad \textbf{Silvio Amir}\quad\quad \textbf{Byron C. Wallace} \\ 
Northeastern University  \\ 
\texttt{\{wadhwa.s, s.amir, b.wallace\}@northeastern.edu} 
}

\begin{document}
\maketitle
\begin{abstract}
Eliciting ``chain of thought'' (CoT) rationales---sequences of token that convey a ``reasoning'' process---has been shown to consistently improve LLM performance on tasks like question answering. More recent efforts have shown that such rationales can also be used for \emph{model distillation}: Including CoT sequences (elicited from a large ``teacher'' model) in addition to target labels when fine-tuning a small student model yields (often substantial) improvements. In this work we ask: {\bf Why and how does this additional training signal help in model distillation?} We perform ablations to interrogate this, and report some potentially surprising results. Specifically: (1) Placing CoT sequences \emph{after} labels (rather than before) realizes consistently better downstream performance---this means that no student ``reasoning'' is necessary at test time to realize gains. (2) When rationales are appended in this way, they need not be coherent reasoning sequences to yield improvements; performance increases are robust to permutations of CoT tokens, for example. In fact, (3) a small number of key tokens are sufficient to achieve improvements equivalent to those observed when full rationales are used in model distillation. 
\end{abstract}

\section{Introduction}
\begin{figure}
\centering
  \includegraphics[scale=0.41]{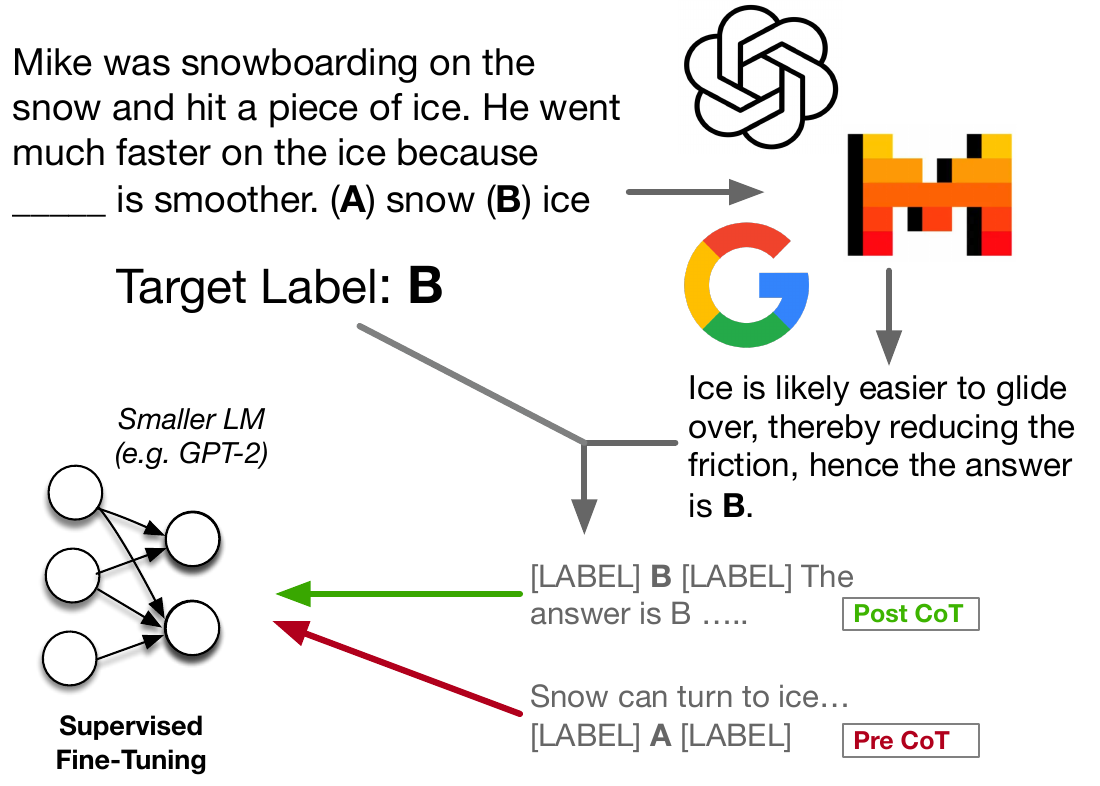}
  \caption{For {\bf RQ1}, we investigate augmenting CoT rationales obtained by very large (teacher) language models like Mistral, \textit{after} the target labels. In doing so, we inject the same CoT reasoning ability during supervised fine-tuning (SFT) but do \textit{not} condition generation of target label on the CoT itself at inference time.}
  \label{fig:r1}
\end{figure}

Chain of thought (CoT) reasoning---i.e., 
generating tokens which communicate step-by-step ``thinking''---can (sometimes dramatically) improve model performance on reasoning tasks \cite{wei2023chainofthought}. 
In the context of \emph{model distillation} \cite{hinton2015distilling}, recent work has elicited 
such rationale chains from massive 
LLMs (e.g., GPT-4) to augment 
data with which to fine-tune much smaller ($<$2B parameters) task-specific models. 
Figure \ref{fig:r1} illustrates this distillation approach: The student model is trained to generate the rationales 
in addition to the target token(s). 

This simple CoT-augmented distillation strategy consistently and sometimes dramatically improves the performance of student models \cite{ho-etal-2023-large}. 
For example, \citet{li-etal-2023-symbolic} used rationales from 
GPT-3 (175B) 
to teach a comparatively tiny student LM (OPT-1.5B) to produce similar ``reasoning''  token sequences at inference time. They show an average increase in task accuracy of 12.4\% across three commonsense reasoning datasets. \citet{shridhar-etal-2023-distilling} adopted a similar approach 
to fine-tune GPT-2 (large; 774M) on grade-school math datasets with improvements 
of 8.23\% on GSM8K and 16.20\% on SVAMP. 
Beyond commonsense reasoning, \citet{wadhwa-etal-2023-revisiting} 
achieved 
SOTA results ($+$6.23 absolute gain in micro-F1, on average) with a distilled model for relation extraction by exploiting CoT rationales. 


In this work we ask: {\bf \emph{Why} does distillation with CoT augmented targets consistently improve the performance of distilled LMs?} 
One might naively suspect that the student model benefits from learning to mimic the relevant ``reasoning'' process. 
But we find that it is not the case that student models benefit from ``reasoning'' at inference time. 

Rather, consistent with contemporaneous work \cite{chen2024postsemanticthinking}, we observe that placing CoT sequences \emph{after} target tokens for distillation actually \emph{improves} student performance (compared to when CoT is pre-prended to labels). 
This means the student model need not  bother generating its ``reasoning'' at test time, as the label will be generated ahead of this anyway.  
Further, we find that rationale grammatically is not necessary; one can shuffle rationale tokens and/or include \emph{only} ``important'' tokens from chains of thought during distillation and still realize performance benefits equivalent to those observed when using the full rationales. 



Through ablations with three small student LMs (GPT-2, Phi-1.5, and Gemma-2B), we report the following, sometimes counter-intuitive, findings regarding CoT-augmented distillation and how rationales benefit student models. 
We summarize our key findings as

\begin{enumerate}
    \item {\bf CoT-augmented distillation works better when rationales are provided \emph{after} labels}. Standard CoT reasoning elicited zero-shot from massive LMs yields rationales as \emph{prefixes} that logically lead to the label token(s). But we find that smaller models perform consistently \textit{better} when rationales \emph{follow} labels in distillation targets.

    \item When \textit{appended} to 
    target labels, {\bf token-level order, length, and coherence of rationales does not matter}. However, 
    these things \emph{do} matter when rationales are preprended. 
    When the rationales are placed \textit{before} the final label during fine-tuning, masking, shuffling, or altering coherent rationales significantly degrades model performance.

    
    \item 
    Motivated by the preceding observations, 
    we run controlled experiments to establish that {\bf there are certain key, contextual tokens that connect the input to the final label, and appending these tokens to labels is sufficient to achieve performance on-par with coherent CoT-like rationales}. It is solely the presence of these tokens at training time that leads to downstream performance improvements. 

\end{enumerate}

\section{Experimental Design}
\label{sec:exp_design}
CoT-augmented distillation entails eliciting rationales from a large \emph{teacher} model and using these as additional training signal for a small \emph{student} model.
Rationales here comprise
the logical steps 
taken to reach a response from a given input.\footnote{In the case of distillation, where one has access to reference labels, one can elicit rationales from the teacher which support the \emph{correct} answer.} 
These are inserted into distillation training targets, and the student model is in this way taught to generate reasoning in addition to labels. 

This has been shown empirically to provide (sometimes dramatic) performance benefits \cite{li-etal-2023-symbolic}. 
But why? What accounts for the success of CoT-augmented distillation? 
In this work we investigate 
the following questions about the role of CoT-rationales in distillation. 
({\bf RQ1}) Does the placement of the reasoning chain relative to the target label (pre- or post-) matter? Relatedly, might observed performance gains owe to simply allowing the student model additional compute during inference? 
({\bf RQ2}) Must rationales feature logical and coherent ``chain-of-thought'' reasoning, or could we, e.g., scramble the ordering of tokens and still observe improvements? 
Finally, ({\bf RQ3}) could we realize the same benefits in distillation using only a handful of key tokens from rationales, rather complete reasoning sequences? 

To answer these questions empirically, 
we establish baseline student LM performance, and then compare this to ablated variants of CoT-augmented models.  
We use a fixed ICL prompt (Appendix \ref{appx:prompts}) with the input and target label to elicit a possible rationale for each instance in a dataset. 
We use Mistral-7B-Instruct \cite{jiang2023mistral} as the teacher model and GPT-2~\cite{radford2019language}, Gemma-2B~\cite{gemmateam2024gemma} and Phi-1.5~\cite{li2023textbooks} as student models. 
Note that one could instead replace Mistral-7B-Instruct with GPT-4 (or any other LLM capable of generating CoT-style rationales in ICL 
settings) as the teacher model. 
See Appendix \ref{appx:prompts} for the prompt used to elicit rationales for training instances of all datasets used in our work. 


Following prior related work \cite{wei2023chainofthought, li-etal-2023-symbolic}, 
we select three commonsense reasoning datasets: CommonsenseQA \cite{talmor-etal-2019-commonsenseqa}, OpenbookQA \cite{mihaylov-etal-2018-suit}, and QuaRel \cite{Tafjord2018QuaRelAD}. 
Each dataset provides an input 
consisting of a question, and a predefined set of answer choices. 
The target labels 
are the correct answer choices (Appendix \ref{appx:datasets}).

\paragraph{Implementation details} We performed all of our experiments on two NVIDIA A100 GPUs. 
All student models (including ablations) were fine-tuned with a learning rate of $3e$-$5$, batch size of 4 for CommonsenseQA and OpenBookQa, and 8 for QuaRel, with a maximum input length of 512, maximum output length of 256. 
We evaluated checkpoints every 500 steps with early stopping (patience $=10$, threshold $=0.02$). 
Because we are only interested in measuring relative performance of fine-tuned models across ablations (as opposed to necessarily realizing SOTA performance), we left the remaining hyperparamters to their default values.


\begin{table}[]
\centering
\small
\begin{tabular}{@{}llrrr@{}}
\cmidrule(l){3-5}
                                                                                &          & \multicolumn{1}{l}{CSQA} & \multicolumn{1}{l}{OBQA} & \multicolumn{1}{l}{QuaRel} \\ \midrule
\multirow{3}{*}{\begin{tabular}[c]{@{}l@{}}Baseline\\ (w/o CoT)\end{tabular}}   & GPT-2    & 63.11                    & 60.20                          & 59.05                      \\
                                                                                & Phi-1.5  & 67.77                    & 56.81                          & 76.82                      \\
                                                                                & Gemma-2B & 68.53                       & 58.15                             & 73.39                         \\ \midrule
\multirow{3}{*}{\begin{tabular}[c]{@{}l@{}}CoT \\ before\\ Label\end{tabular}} & GPT-2    & 67.20                    & 69.71                          & 66.27                      \\
                                                                                & Phi-1.5  & 70.83                    & 63.49                          & 79.99                      \\
                                                                                & Gemma-2B & 70.61                       & 65.85                             & 74.90                         \\ \midrule
\multirow{3}{*}{\begin{tabular}[c]{@{}l@{}}CoT\\ after\\ Label\end{tabular}}   & GPT-2    & 70.92                    & 70.26                          & 71.04                      \\
                                                                                & Phi-1.5  & 72.56                    & 72.49                          & 81.36                      \\
                                                                                & Gemma-2B & 72.64                       & 68.93                             & 78.16                         \\ \bottomrule
\end{tabular}
\caption{Comparison of decoder-only models' performance under baseline supervised fine-tuning (no CoT), standard (pre) CoT, and postfix CoT. }
\label{tab:position}
\end{table}

\begin{figure*}
\centering
  \includegraphics[scale=0.044]{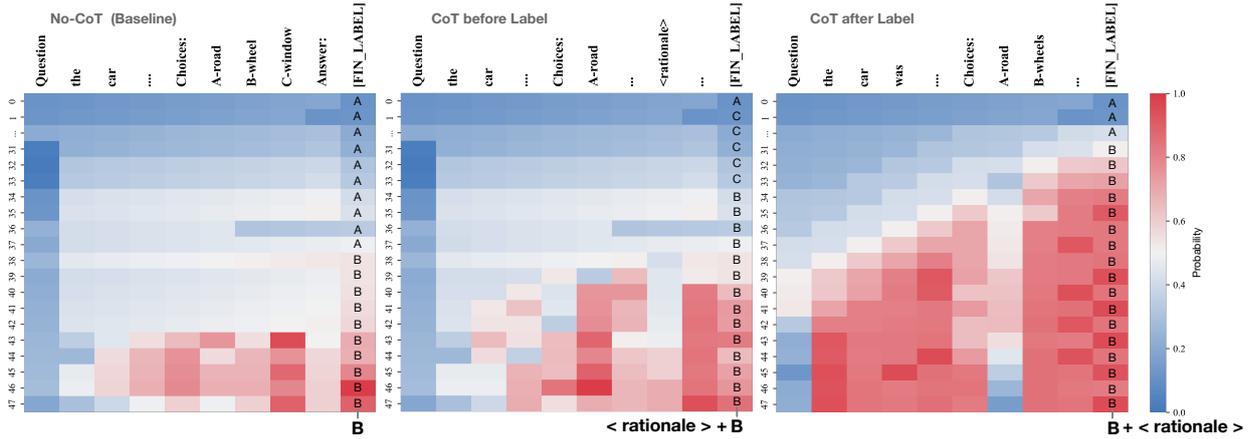}
  \caption{TunedLens \cite{belrose2023eliciting} visualizations on GPT-2 variants fine-tuned without CoT rationales (left), and with them pre-pended (middle) and appended (right). Augmenting distillation with CoT results in models that are more confident in labels earlier on. Models trained with rationales following labels are especially confident.} 
  \label{fig:logitlens}
\end{figure*}

\section*{RQ1: Positioning of Rationales}

Does it matter if we place CoT rationales before or after target labels prior to distillation? 
Prior 
work \cite{wei2023chainofthought} which elicited CoT reasoning from 
LLMs at inference time found that generating the chain \textit{after} the final label performs comparably to the baseline (i.e., no CoT).
This would seem to suggest that ``reasoning'' at inference time is what yields improvements, but it is unclear whether this 
holds in the context of model distillation. 
 We compare the performance of student LMs distilled from examples with rationales placed both \emph{before} and \emph{after} labels
(Figure \ref{fig:r1}). 

\begin{flushleft}
\texttt{\textbf{CoT before Label} Friction is higher on rougher....[FIN\_LABEL] \textbf{B} [FIN\_LABEL]}

\texttt{\textbf{CoT after Label} [FIN\_LABEL] \textbf{B} [FIN\_LABEL] Friction is higher on rougher....}
\end{flushleft}


We find that generating the CoT rationale \textit{before} the label under-performs generating the rationale \textit{after} the label. These findings are consistent across models and datasets (Table \ref{tab:position}).
Note that models trained to generate a CoT after 
the target label, do not need to do so at inference time.
While in general CoT elicited from massive models is thought to improve performance by enabling explicit reasoning, gains offered in the context of distillation must be realized via some other mechanism (e.g., enriched training signal).  

Next, 
we 
examine how and with what confidence 
do models fine-tuned under different conditions 
encode label information. 
We use ideas from LogitLens \cite{logit_lens}, TunedLens \cite{belrose2023eliciting}, and FutureLens \cite{Pal_2023}, which suggest that decoder-only models 
``think iteratively'' and can be probed by inducing a distribution over the output vocabulary conditioned on hidden states to measure model confidence at different layers and time-steps within the model. 

For each dataset, we look at test instances that are \textit{correctly} predicted by all three model types, i.e., models distilled using: 
(i) No CoT; (ii) CoT before label; and (iii) CoT after label.

Figure \ref{fig:logitlens} illustrates 
model confidences (i.e., probabilities computed with a softmax over the LM-head predictions for the final label) at different layers and time points, 
up to and including the final label prediction.\footnote{Full outputs omitted for brevity. See Appendix \ref{appx:lensviz} for full length heatmaps.}

In $80\%$ of correctly predicted outputs, models trained with rationales \textit{appended} to the final label (right-most subplot, Figure \ref{fig:logitlens}) correctly predict the label with probability $>0.6$ at layer 32 and above. 
By contrast, models trained without any rationales (left-most subplot) lack such confidence, especially at lower layers: 
Final label probability does not exceed $0.6$ until layer 44. 
Finally, for models trained with rationales \textit{prepended} to target labels (middle sub-plot), the probability of the true label is $\leq$ $0.6$ until layer 39 in $80\%$ of correctly predicted instances. 
In sum, this analysis (illustrated 
in Figure \ref{fig:logitlens}) reveals a clear difference: Added CoT-information during distillation yields models which are more confident earlier on (positionally and layerwise) in the final output. 

\begin{figure}[t]
\centering
  \includegraphics[scale=0.455]{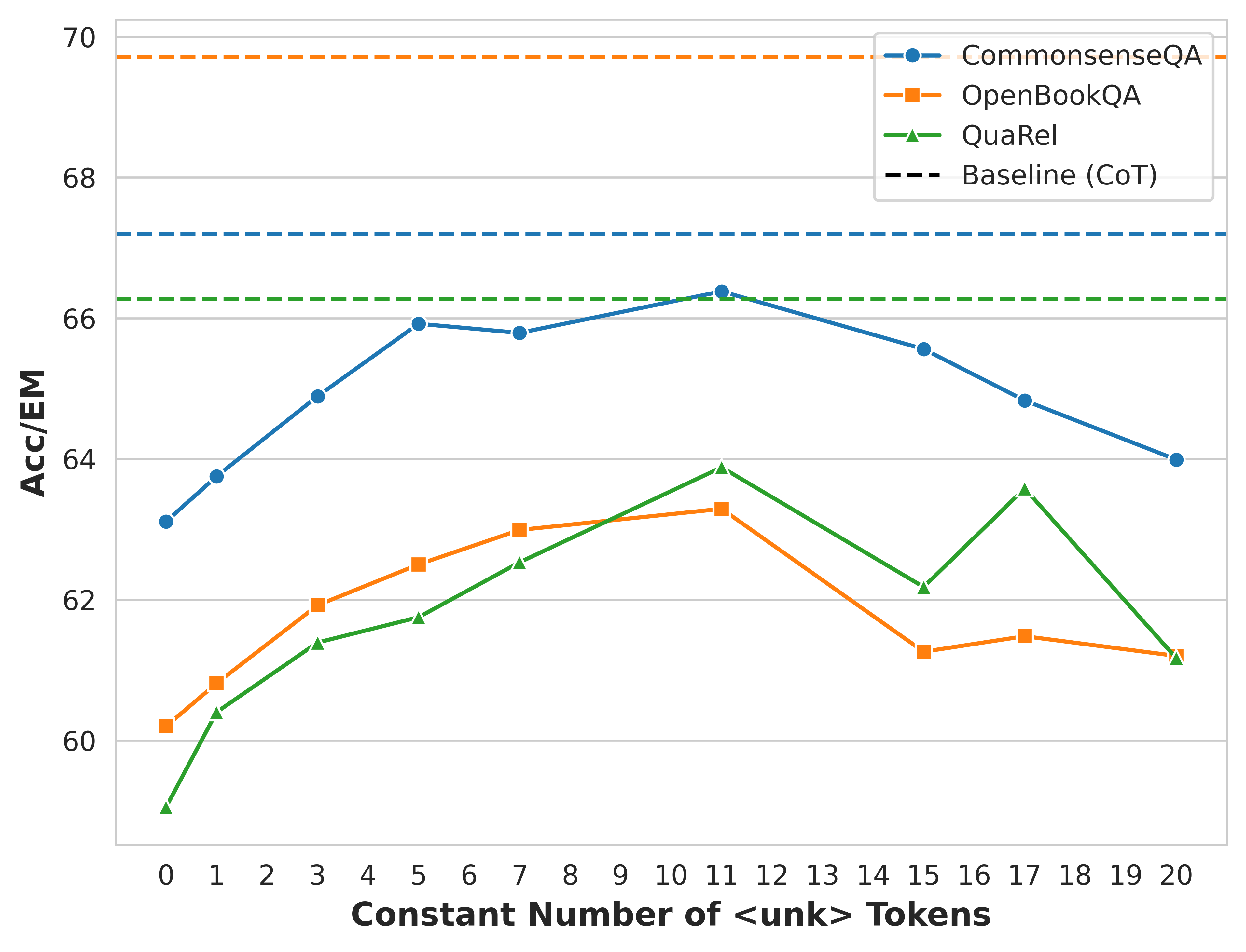}
  \caption{Performance of GPT-2 with constant number of $<${\tt unk}$>$ tokens prepended to the target label.}
  \label{fig:r2_unks}
\end{figure}

\paragraph{Is it \textit{just} the extra ``compute''?} 
Prior work by \citet{goyal2024think} observed performance improvements in LLMs when inputs were augmented with ``dummy'' tokens (at pretraining and inference time), suggesting that LLMs benefit from additional compute cycles. Here we 
investigate whether it is just the added compute (i.e., steps/gradient updates over target label during fine-tuning) that provides gains \textit{comparable to those achieved with CoT}, or if it is the CoT rationales contain useful information. Instead of CoT rationales, we prepend a fix-sized sequence of $<${\tt unk}$>$ tokens to the target label and ablate over the sequence length. 

\begin{figure*}[]
\centering
  \includegraphics[width=6.3in]{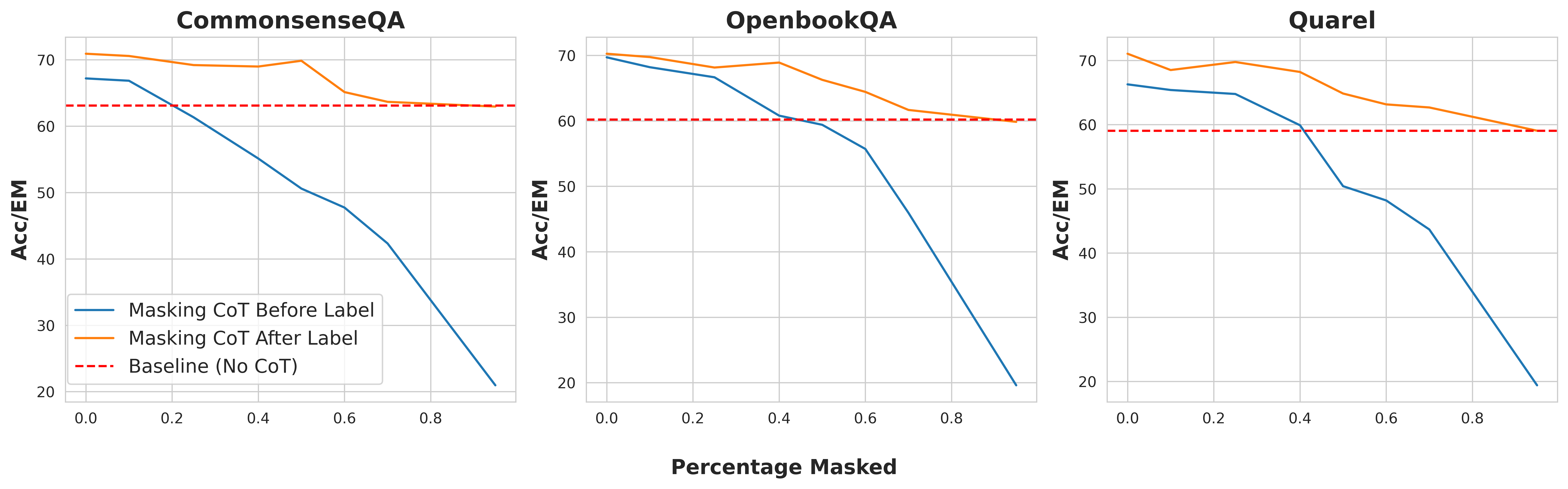}
  \caption{Comparison of model performance while successively reducing the amount of available information in a CoT rationale through masking.}
  \label{fig:masking}
\end{figure*}

Figure \ref{fig:masking} summarizes our results with GPT-2 as the student model; similar to \citet{goyal2024think} we observe that adding compute steps 
during training leads to 
(sometimes substantial) improvements in downstream performance. 
However, beyond a certain point ($\sim11$ $<$unk$>$ tokens) performance plateaus, and then eventually declines. 
More importantly, at no point does the model outperform a CoT baseline (Table \ref{tab:position}), suggesting that CoT rationales do indeed incorporate information necessary to achieve downstream improvements.

\begin{table*}[]
\centering
\small
\begin{tabular}{@{}llrrrllrrr@{}}
\cmidrule(lr){3-5} \cmidrule(l){8-10}
\multicolumn{1}{c}{}                                                                 & \multicolumn{1}{c}{} & CSQA  & OBQA  & QuaRel                     &                                                                                     &          & \multicolumn{1}{l}{CSQA} & \multicolumn{1}{l}{OBQA} & \multicolumn{1}{l}{QuaRel} \\ \midrule
\multirow{3}{*}{\begin{tabular}[c]{@{}l@{}}CoT\\ before Label\end{tabular}}                                                    & GPT-2                & 67.20 & 69.71 & \multicolumn{1}{r|}{66.27} & \multirow{3}{*}{\begin{tabular}[c]{@{}l@{}}CoT\\ after Label\end{tabular}}                                                     & GPT-2    & 70.92                    & 70.26                    & 71.04                      \\
                                                                                     & Phi-1.5              & 70.83 & 63.49 & \multicolumn{1}{r|}{79.99} &                                                                                     & Phi-1.5  & 72.56                    & 72.49                    & 81.36                      \\
                                                                                     & Gemma-2B             & 70.61    & 65.85    & \multicolumn{1}{r|}{74.90}    &                                                                                     & Gemma-2B & 72.64                       & 68.93                       & 78.16                         \\ \midrule
\multirow{3}{*}{\begin{tabular}[c]{@{}l@{}}Shuffled CoT\\ before Label\end{tabular}} & GPT-2                & 34.56 & 41.64 & \multicolumn{1}{r|}{32.88} & \multirow{3}{*}{\begin{tabular}[c]{@{}l@{}}Shuffled CoT\\ after Label\end{tabular}} & GPT-2    & 69.56                    & 70.15                    & 70.56                      \\
                                                                                     & Phi-1.5              & 19.37 & 38.81 & \multicolumn{1}{r|}{45.28} &                                                                                     & Phi-1.5  & 72.19                    & 69.51                    & 81.01                      \\
                                                                                     & Gemma-2B             & 25.80    & 35.17    & \multicolumn{1}{r|}{20.52}    &                                                                                     & Gemma-2B & 71.13                       & 67.28                       & 76.50                         \\ \bottomrule
\end{tabular}
\caption{Comparison of model performance when shuffling the rationales to test for robustness to CoT coherence during SFT.}
\label{tab:shuffling}
\end{table*}

\section*{RQ2: Tokens within CoT Rationales} 
\label{sec:rq2}
In light of our findings from RQ1, we 
next investigate what specific information in CoT rationales 
improves downstream 
performance. 
To this end, we assess how robust student models are to perturbations of provided rationales. 
Specifically, we consider: (i) Shuffling tokens within a rationale; and (ii) Incrementally masking tokens while retaining their relative order. 

\paragraph{Shuffling} We start by testing the robustness of student LMs with respect to the \emph{coherence} of rationales.  
In particular, we shuffle tokens comprising rationales at the instance level. 
To illustrate this, consider the following example. 

\begin{flushleft}
\texttt{\textbf{Question}: If you hired a pitcher, (A) a nerd (B) a bodybuilder, who likely can pitch a baseball faster?} 

\texttt{\textbf{Original CoT Rationale}: The answer is B because bodybuilders typically have more strength than nerds, which could translate into a greater ability to throw a baseball faster. [FIN\_LABEL] B [FIN\_LABEL]}

\texttt{\textbf{Shuffled Rationale}: Baseball a faster throw to ability greater a into translate could which nerds than strength more have typically bodybuilders because B is answer The. [FIN\_LABEL] B [FIN\_LABEL]}
\end{flushleft}

We then train the student LM with these shuffled rationales in place of the original (coherent) versions, under both pre- and post-label settings.  
Table \ref{tab:shuffling} summarizes our findings from these experiments. 
We see that \emph{prepending} the shuffled CoT rationales to target labels leads to sharp decline in performance, whereas \emph{appending} these to target labels has nearly no effect on subsequent model performance. 

Taken together with the results from RQ1, we hypothesize that 
this may be because prepending rationales to target labels during distillation requires the student model to learn to generate coherent rationales in addition to producing correct labels.
By contrast, when rationales are appended they can serve as additional supervision during training without requiring coherent rationale generation at inference time. 

\paragraph{Masking} Next we run an ablation intended to test whether 
the full rationales are needed or if a subset of words is sufficient to realize the observed benefits.
We start by randomly masking varying fractions of tokens within a rationale. 
For example: 

\begin{flushleft}
\texttt{\textbf{Question}: If you hired a pitcher, (A) a nerd (B) a bodybuilder, who likely can pitch a baseball faster?} 

\texttt{\textbf{Original CoT Rationale}: The answer is B because bodybuilders typically have more strength than nerds, which could translate into a greater ability to throw a baseball faster. [FIN\_LABEL] B [FIN\_LABEL]}

\texttt{\textbf{10\% Masked}: The answer is [MASK] because bodybuilders typically have more [MASK] than nerds, which could translate into a greater ability to throw a baseball [MASK]. [FIN\_LABEL] B [FIN\_LABEL]}

\texttt{\textbf{50\% Masked}: The [MASK] is B [MASK] bodybuilders typically [MASK] more [MASK] than nerds, [MASK] could translate [MASK] a greater [MASK] to throw [MASK] baseball [MASK]. [FIN\_LABEL] B [FIN\_LABEL]}

\texttt{\textbf{90\% Masked}: [MASK] [MASK] [MASK] B [MASK] [MASK] [MASK] [MASK] [MASK] [MASK] [MASK] 
[MASK], [MASK] [MASK] [MASK] [MASK] [MASK] [MASK] [MASK] [MASK] [MASK] [MASK] [MASK] [MASK] [MASK] 
[MASK] [MASK] [MASK] [MASK] [MASK] faster. [FIN\_LABEL] B [FIN\_LABEL]} 
\end{flushleft}

\noindent We vary the proportion 
of masked tokens from 10\% to 90\% (in increments of 10-15\%) and again test under 
both pre- and post-label settings (see RQ1).

Figure \ref{fig:masking} reports performances as a function of 
the proportion of masked tokens as compared to a non-CoT baseline. 
When only a small fraction (up to $\sim$20\%) of rationale tokens are masked, we observe only marginal performance declines in both 
settings. 
However, as the proportion 
of masked CoT tokens increases (40\%$+$), we see rapid performance decline in the CoT before label case ---
at 60\% masking, we find that 
the resultant distilled model performs worse than the baseline (i.e., without CoT). 

We observe that masking a high percentage of tokens prior to the label 
yields models that generate a variable but often large number of [{\tt MASK}] tokens prior to target label, often reaching maximum output length (set as a decoding hyperparameter).  
In contrast, in the CoT after label 
setting we observe gains over the non-CoT distillation baseline up until a high fraction ($>60$\%) of tokens are masked; and 
beyond this point, the performance matches the vanilla (non-CoT) baseline. 

 

\begin{table}[]
\centering
\small
\begin{tabular}{@{}llrrr@{}}
\cmidrule(l){3-5}
                                                                                 &          & 
                                                                                 \multicolumn{1}{l}{CSQA} & \multicolumn{1}{l}{OBQA} & \multicolumn{1}{l}{QuaRel} \\ \midrule
\multirow{3}{*}{\begin{tabular}[c]{@{}l@{}}Baseline\\ (no CoT)\end{tabular}}   & GPT-2    & 63.11                    & 60.20                          & 59.05                      \\
                                                                                & Phi-1.5  & 67.77                    & 56.81                          & 76.82                      \\
                                                                                & Gemma-2B & 68.53                       & 58.15                             & 73.39                         \\ \midrule
\multirow{3}{*}{\begin{tabular}[c]{@{}l@{}}CoT \\ after \\ Labels\end{tabular}}  & GPT-2    & 70.92                    & 70.26                          & 71.04                      \\
                                                                                 & Phi-1.5  & 72.56                    & 72.49                          & 81.36                      \\
                                                                                 & Gemma-2B & 72.64                       & 68.93                             & 78.16                         \\ \midrule
\multirow{3}{*}{\begin{tabular}[c]{@{}l@{}}Grad \\ Attr\end{tabular}}            & GPT-2    & 71.30                    & 74.86                          & 71.26                      \\
                                                                                 & Phi-1.5  & 74.82                    & 71.54                          & 82.69                      \\
                                                                                 & Gemma-2B & 73.85                       & 68.13                             & 79.03                         \\ \midrule
\multirow{3}{*}{\begin{tabular}[c]{@{}l@{}}Grad \\ Attr\\ Shuffled\end{tabular}} & GPT-2    & 71.24                    & 74.99                          & 71.47                      \\
                                                                                 & Phi-1.5  & 74.18                    & 71.28                          & 81.84                      \\
                                                                                 & Gemma-2B & 72.93                       & 67.30                             & 78.94 \\ \midrule
\multirow{3}{*}{\begin{tabular}[c]{@{}l@{}}Human\\ Labels\end{tabular}}            & GPT-2                      & --    & --    & 67.06  \\
                                                                                   & Phi-1.5                    & --    & --    & 78.44  \\
                                                                                   & Gemma-2B                   & --    & --    & 74.77     
                                                                                 \\ \midrule
\multirow{3}{*}{\begin{tabular}[c]{@{}l@{}}Word2Vec\\ Based\end{tabular}}            & GPT-2                      & 63.81    & 60.02    & 59.90  \\
                                                                                   & Phi-1.5                    & 67.94    & 56.22    & 75.49  \\
                                                                                   & Gemma-2B                   & 69.10    & 58.86    & 72.12     
                                                                                 \\
                                                                                 \bottomrule
\end{tabular}
\caption{Comparison of model performance under different attribution methods relative to retaining full length post-label CoT rationales.}
\label{tab:grad_attr}
\end{table}

\section*{RQ3: Attribution from Rationales}
\label{sec:r3}
Having established that 
placing rationales 
after labels yields the best performance 
when performing CoT-augmented distillation---even without the full reasoning chain---we now ask whether we can find a small subset of ``important’’ tokens that are sufficient to realize performance benefits.
To determine importance, we consider both gradient-based attribution and human annotations. 

\begin{figure*}
\centering
  \includegraphics[scale=0.12]{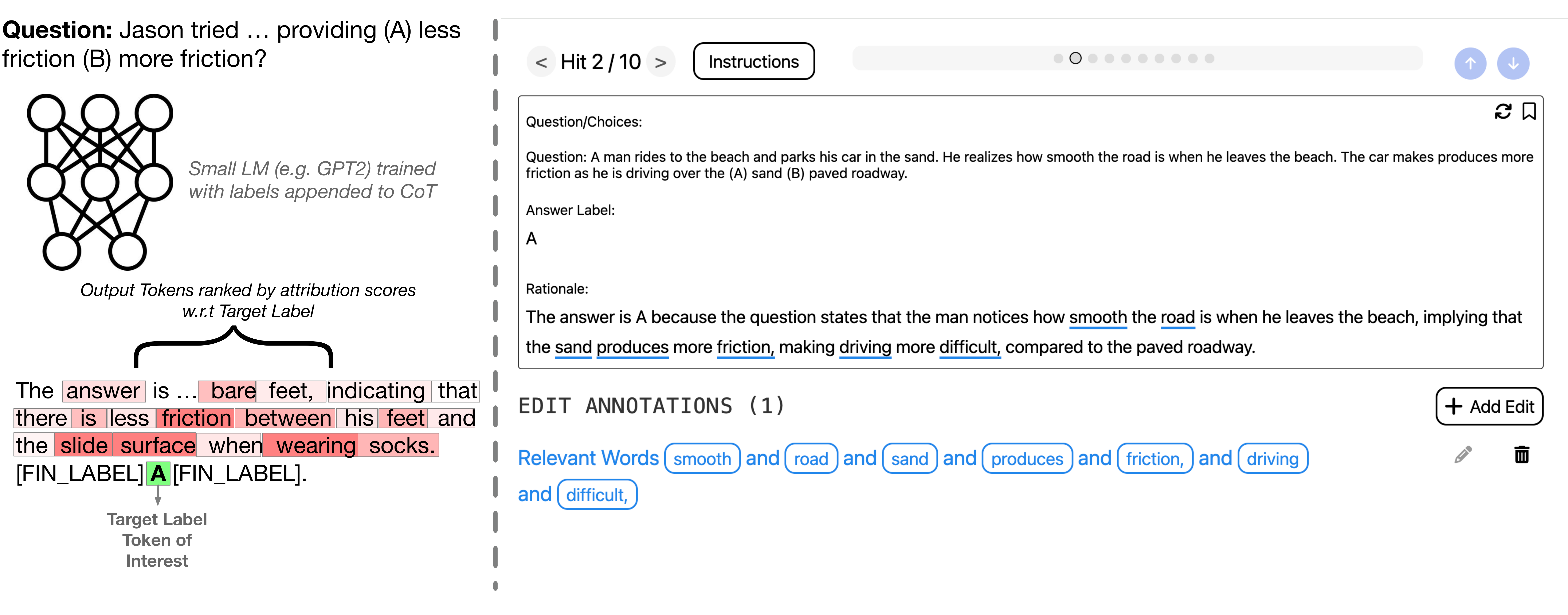}
  \caption{Comparison of Attribution Methods: Left side we have automated extraction via Integrated Gradients while the right side displays manually annotated words \textit{perceived} by human annotators to be relevant.}
  \label{fig:gradattr}
\end{figure*}

\paragraph{Attribution via integrated gradients} is a method to estimate the importance of individual tokens with respect to model output~\cite{Sundararajan2017}. 
To measure the relative importance of rationale tokens on the final target label, we start with a baseline model (GPT-2) which is fine-tuned to generate a CoT-rationale \textit{before} the final label. Considering a token sequence \( \mathbf{x} = [x_1, x_2, \ldots, x_m, \ldots, x_n] \) where $x_{1, \ldots, m}$, correspond to the input tokens;  $x_{m+1, \ldots,n-1}$ are the rationale tokens; and $x_{n}$ is the final target label, we compute an approximation of the integrated gradients for the \( i \)-th rationale token as: 



\begin{align*}
\mathrm{IG}_i &\approx (x_i - x'_i) \sum_{k=1}^{p} \frac{\partial f(\mathbf{x'} + \frac{k}{m} (\mathbf{x} - \mathbf{x'}))}{\partial x_i} \cdot \frac{1}{p}
\end{align*}

\noindent where, \( \mathbf{x'} \) is a baseline (zero vector or a neutral input), \( f(\mathbf{x}) \) represents the model’s output and \( p \) is the granularity for the approximation. 
See \citealt{Sundararajan2017} 
for details. 


The average length of a CoT rationale in our data is 36.3 tokens and 
we retain the top 15 tokens with the highest attribution scores. Figure \ref{fig:gradattr} illustrates the process of computing a set of important tokens for the target label. As an example from QuaRel's training set\footnote{Training instance ID: V1\_B3\_0128}--  
\\
\begin{flushleft}
\texttt{\textbf{Question}: If you hired a pitcher, (A) a nerd (B) a bodybuilder, who likely can pitch a baseball faster?} 

\texttt{\textbf{Original CoT Rationale}: [FIN\_LABEL] B [FIN\_LABEL] The answer is B because bodybuilders typically have more strength than nerds, which could translate into a greater ability to throw a baseball faster.}

\texttt{\textbf{Attributed Tokens}: [FIN\_LABEL] B [FIN\_LABEL] B because body builders more strength translate throw baseball faster}

\texttt{\textbf{Shuffled Attributed Tokens}:[FIN\_LABEL] B [FIN\_LABEL] translate because more body B faster builders baseball strength throw}
\end{flushleft}

We fine-tune the models again, 
\textit{appending} the attributed tokens to the final label, similar to the case where CoT rationales are generated after the label at inference. Table \ref{tab:grad_attr} summarizes our findings. We broadly observe that reducing the total number of tokens in CoT rationale via gradient attribution leads to no significant difference in downstream model performance. This is consistent with our findings in RQ2, where masking a majority of CoT tokens \textit{appended} to the target label did not significantly effect model performance. 

\paragraph{Human annotations} As an alternative to 
scoring rationale tokens via gradient attribution, 
we evaluate using tokens that humans \textit{perceive} to be the most 
relevant to target labels. 
We hire annotators on  Prolific\footnote{\url{https://www.prolific.com/}} to identify minimal sets of (up to 15) words in rationales necessary to answer the question (Figure \ref{fig:gradattr}).\footnote{Interface designed using \url{https://thresh.tools}.}  
We first ran a small internal pilot to estimate time required 
and set fair pay rates.\footnote{We pay US\$15/hr to all crowdworkers regardless of their geographic location.}
Next we 
collected annotations for $\sim$2k instances of the QuaRel dataset in batches of 200
from crowdworkers fluent in English. 
We manually verified $10\%$ (20 instances) of each batch to ensure quality.\footnote{We required all crowdworkers to have an overall job approval rating of $\ge$95$\%$ with at least 100 completed jobs on Prolific.} 

Replacing CoT rationales with words 
deemed important by annotators offers some gains, 
but smaller than those from integrated gradients 
(Table \ref{tab:grad_attr}). 
To 
measure overlap between tokens selected manually and via gradient attribution, 
we assume the latter to be the reference tokens and measure Precision ($0.73$) and Recall ($0.59$) of annotated words\footnote{We assume a complete overlap if any subword in an annotated word matches with a reference token.} 
In general, we find 
the set of tokens identified through gradient attribution to be much more comprehensive than those selected by human annotators. 


\paragraph{Are tokens ``similar to'' the label sufficient?} Finally, we explore whether CoT rationales elicited from LLMs merely provide 
tokens that are similar to but distinct from target 
labels. 
One way to collect such tokens is to  select from a large set of words a subset that have high similarity to the target label token. 
To this end we use static  Word2Vec \cite{mikolov2013distributed} embeddings.\footnote{\texttt{word2vec-google-news-300}\footnote{\url{https://radimrehurek.com/gensim/auto_examples/tutorials/run_word2vec.html}}; trained on the Google News dataset of $\sim$100 billion words.}

We select the 15 closest words to the target label for all training instances. 
For target labels with multiple tokens, we take similarity with respect to only the longest token. 
We then use these retrieved tokens in lieu of CoT rationales when fine-tuning student models. 
The question is whether the additional information encoded in words similar to the target label yield performance gains comparable to CoT augmented distillation. 
Table \ref{tab:grad_attr} reports results, which are largely negative for this experiment: The observed performance with ``relevant word augmentation'' is comparable to the no-CoT setting (baseline). 
This suggests that while rationales need not be coherent to realize benefits, the tokens they comprise must offer additional signal beyond being simply ``similar'' to (in terms of co-occurence) target label tokens. 

\section{Related Work}
\paragraph{Distillation via elicitation}
\citet{west-etal-2022-symbolic} considered ``symbolic'' distillation where instead of distilling from soft representations like logits, they proposed the use of LLMs as data generators to be used to augment training data. 
Other recent work has shown that \textit{explanations} can serve as both inputs \cite{hase-bansal-2022-models} and targets \cite{wiegreffe-etal-2022-reframing}, and can be used downstream to improve task- \cite{wadhwa-etal-2023-revisiting} and domain-specific \cite{ho-etal-2023-large} model performance.

\citet{li-etal-2023-symbolic} first explored distillation performance to tasks like commonsense reasoning and provided analyses intended to reveal factors that may be important in creating the \textit{teacher} corpus, upon which our work builds on. 
Beyond directly using explanations-style rationales for fine-tuning, \citet{deng2023implicit} explored an alternative approach by using model hidden states to perform \textit{implicit} reasoning, instead of producing rationale tokens one-by-one (i.e. Next Token Prediction), demonstrating that the chains of thought themselves may not be fully necessary to achieve downstream fine-tuning performance gains. 

Our work deepens these  efforts by focusing on analyzing specific fine-tuning for distillation dynamics in smaller models, and characterizing when rationales generated by teacher LLMs are helpful.

\paragraph{CoT with Small Models}
In-context CoT prompting \cite{wei2023chainofthought} induces thinking \textit{step-by-step}, such that the model generates intermediate reasoning ultimately leading to a target label. 
Prior work \cite{ho-etal-2023-large,magister2023teaching} has shown that small models may not be not inherently capable of generating these reasoning chains, but can be taught to do so using augmented training sets. 

Creating CoT-augmented training sets can be expensive, and a number of prior works in the area have investigated synthetic data generation. For instance, \citet{Hsieh2023DistillingSO} generate new target labels from few instances of labeled data. \citet{li-etal-2023-symbolic} notably found that sampling multiple rationales can improve small-model performance. \citet{han2023dialcot} decomposed the reasoning steps into multi-round dialog and optimize for the correct path using PPO algorithm while training smaller models. \citet{Fu2023SpecializingSL} emphasize the trade-offs between task-specific CoT-generation capability in small models and their generalizability. \citet{wang-etal-2023-scott} establish the effect of faithfulness of elicited rationales on the student models trained using them. A shared theme in these past papers have been that they explicitly look at improving the \textit{quality} of rationales themselves and its downstream effects on overall model performance.

Our work differs from these efforts looking only at the final label, manipulating the CoT rationales \textit{at distillation (fine-tuning) time} to probe how rationales effect model performance.  

\paragraph{Contemporaneous Work}
While engaged in this work, a few contemporaneous 
efforts have surfaced which make some observations that overlap with our findings. 
\citet{chen2024postsemanticthinking} introduce ``post-semantic thinking'' (PST) to reduce the influence of 
rationales on final output labels. 
\citet{xu2024preemptive} reveal that preemptive answer generation (a target label) 
within a CoT rationale is highly sensitive to malicious attacks, which comports with our hypothesis (i.e. vice versa) that a faulty reasoning leading to incorrect rationales can effect the overall model performance (which is solely evaluated on labels generated after those rationales). 


\section{Conclusions}

We have investigated \textit{why} and \textit{under what circumstances} does  CoT-augmented 
distillation improve student model performance. 
Specifically, we evaluated the degree to which the following aspects contribute to the observed gains realized in CoT-augmented distillation. 

\begin{enumerate}

\item The placement of rationales (before or after labels). {\bf Finding: Appending (rather than pre-pending) rationales to targets yields consistently better performance.}
\item The coherence of rationales and their grammatically.
{\bf Finding: When rationales follow labels, the words they comprise can be scrambled and one still observes comparable gains.}
\item Whether we need only a small set of key tokens from rationales (and how to identify them). 
{\bf Finding: Gains comparable to CoT-augmented distillation can be realized using a small set of tokens identified via gradient attribution; using manually selected ``important'' words does not do as well, nor does using tokens that are ``similar to'' label words.}
\end{enumerate}

Some of these findings corroborate and deepen observations made in contemporaneous work, e.g., models can benefit from additional compute at inference time \cite{goyal2024think},  
and CoT-augmentation fares best when rationales are placed \emph{after} the target labels \cite{chen2024postsemanticthinking}. 
We have not fully characterized the mechanism by which CoT augmentation aids distillation, but we have ruled out some explanations and provided empirical insights into when and how CoT augmentation provides useful signal to student models.

\section*{Limitations}


There are important limitations to this work and the conclusions we can draw from it. 

First, 
we have only considered publicly available open-domain question-answering datasets in our analyses, to the exclusion of complex information extraction tasks such as relation extraction where CoT-augmented distillation has also proven useful \cite{wadhwa-etal-2023-revisiting}. 
We made this choice largely in the interest of consistency with prior work, and to avoid complex evaluation challenges that occur during generative relation extraction. 

Second, we did not attempt to \textit{improve} the quality of CoT rationales generated by teacher models through iterative prompt refinement or other techniques \cite{wang-etal-2023-scott}. 
We also elicited the rationales for distillation from modestly sized open source models, rather than (for example) GPT-4. 
It may be possible to elicit ``better'' rationales from massive proprietary models, but it seems unlikely (though possible) that our conclusions vis-a-vis distillation would change as a result. 

Third, our evaluation of using rationale tokens annotated manually is limited by the way we framed the task. 
It could be that an alternative design and/or annotation interface would yield different annotations, and this may in turn lead to different conclusions regarding the utility of tokens selected in this way. 

Finally, we \textit{only} experimented with English-language datasets and we therefore cannot say whether these results would hold in other languages. 

\section*{Ethics Statement}
This work required some human annotations which were collected 
using an online platform called Prolific. Prolific required us to pay workers \textit{per hour}, and so we had to estimate the time required to complete one batch of annotations. 
To do so, we (the authors) carried out a small number of these annotations to determine the approximate hourly compensation. We then set the compensation rate to average \$15 USD/hour. 
If annotators took longer than expected to complete a batch of annotations, we paid bonuses to ensure that their cumulative pay averaged out to US\$15/hour.
\paragraph{Statement of Intended Use} Our work relies 
on open source datasets and models. 
Like any trained model, there is a risk of the distilled model inheriting or amplifying any biases present in the original LLM's rationales. While rationales make the model more interpretable than a blackbox classifier, there still may be challenges in fully explaining the distilled model's behavior. While distilling, the user must be aware of these considerations and institute appropriate safeguards. 

\section*{Acknowledgements}

This work was supported in part by the National Science Foundation (NSF) grant IIS-1901117.

\bibliography{anthology,custom}
\clearpage
\appendix
\section*{Appendix}

\section{Dataset Details}
\label{appx:datasets}
We conducted our experiments using three datasets; for completeness we provide details about these here. 

\paragraph{CommonsenseQA} \cite{talmor-etal-2019-commonsenseqa} is a multiple-choice question answering dataset that requires commonsense knowledge. 
Each question is accompanied by five answer choices; only one is correct. 
The dataset consists of 12,102 questions split into a training, development, and test sets of set of 9,741, 1,221, and 1,140 questions, respectively. The following is an example from the training data (ID: \texttt{7e93dacd4d1b7c7aa4c15f5da220bd59})

\begin{flushleft}

\texttt{\textbf{Question:} The two conglomerates decided to reach  tentative agreement to what?}

\texttt{\textbf{Choices:}\\A: do business\\
B: accomplish\\
C: stop arguing\\
D: make progress\\
E: digging holes} 

\texttt{\textbf{Answer:} A (do business)}  


\end{flushleft}

\paragraph{OpenBookQA} \cite{mihaylov-etal-2018-suit} is designed to test an understanding of elementary science, combining factual knowledge with commonsense reasoning. The dataset contains 5,957 questions, each with four answer choices (and one correct response). 
This is split into training, development, and test sets of 4,957, 500, and 500 questions respectively. 
A unique aspect of OpenbookQA is its focus on scientific facts which students are expected to know. The following is an example from the training data (ID: \texttt{12-271})

\begin{flushleft}

\texttt{\textbf{Question:} Skills are learned characteristics. To get better at doing something, you must stretch yourself in ways that}

\texttt{\textbf{Choices:}\\A: may be very uncomfortable at first\\
B: take very little time\\
C: are without learning from others and past experiences\\
D: are without goals and commitment} 

\texttt{\textbf{Answer:} A (may be very uncomfortable at first)}  


\end{flushleft}

\paragraph{QuaRel} \cite{Tafjord2018QuaRelAD} is a dataset for 
reasoning over physical processes involving comparative relationships. It consists of 2,740 multiple-choice questions, each with two answer choices (one being correct). 
The questions require reasoning about how physical processes affect different entities in qualitative ways. 
The dataset provides train/development/test splits comprising 1,948/278/514 questions. The following is an example from the training data (ID: \texttt{QuaRel\_V1\_Fr\_0344})

\begin{flushleft}

\texttt{\textbf{Question:} Ryan races his car and needs to drive in different types of situations. Ryan drives around in a sandy desert, and then in an empty parking lot. After each drive, Ryan sees how warm his car got. Ryan notices that his car was much warmer after driving in the sand than it was after driving in the parking lot. That is because the sand had \_\_\_\_\_ than the parking lot.}

\texttt{\textbf{Choices:}\\A: more resistance\\
B: less resistance} 

\texttt{\textbf{Answer:} A (more resistance)}  

\end{flushleft}

\section{Prompts}
\label{appx:prompts}

Our experiments required eliciting chain-of-thought (CoT) rationales from a ``teacher'' LLM to be used in distillation.
For this we used Mistral-7B-Instruct \cite{jiang2023mistral}. 
We used the following rationale-augmented few-shot prompt to this end.
The question, answer choices, and the target label are taken from the original training instance, and the CoT rationale provided was written by us (the authors). 

\subsection*{CommonsenseQA}
\begin{flushleft}
\small
\texttt{$<$s$>$$[$INST$]$ Given the following two examples of question-answer-rationale triplets, provide a rationale for the third example for why the selected choice answers the question. $[$\textbackslash INST$]$}  

\texttt{\textbf{Question:} The president had to make a decision regarding the hate attack on his country, what did he do?}
\texttt{\textbf{Choices:}A: wage war; B: fight enemy; C: kill; D: destroy enemy; E: attacked his country}

\texttt{\textbf{Answer:} A (wage war)}  

\texttt{\textbf{Rationale:} The answer is A because the president's decision to address a hate attack on his country typically involves taking military action, such as waging war, to protect and defend the nation. $<$/s$>$}

\texttt{\textbf{Question:} Letters are sometimes delivered by hand through one of these?}
\texttt{\textbf{Choices:}A: mail box; B: suitcase; C: front door; D: bowl; E: post office}

\texttt{\textbf{Answer:} C (front door)}  

\texttt{\textbf{Rationale:} The answer is C because letters are delivered by hand through the front door.$<$/s$>$}
\end{flushleft}

\subsection*{OpenBookQA}
\begin{flushleft}
\small
\texttt{$<$s$>$$[$INST$]$ Given the following two examples of question-answer-rationale triplets, provide a rationale for the third example for why the selected choice answers the question. $[$\textbackslash INST$]$}  

\texttt{\textbf{Question:} Oak tree seeds are planted and a sidewalk is paved right next to that spot, until eventually, the tree is tall and the roots must extend past the sidewalk, which means}
\texttt{\textbf{Choices:}A: roots may fall apart; 
B: roots may begin to die; 
C: parts may break the concrete; 
D: roots may be split}; 

\texttt{\textbf{Answer:} C (parts may break the concrete)}  

\texttt{\textbf{Rationale:} The answer is C because as the oak tree grows, its roots may exert pressure on the sidewalk, causing the concrete to crack or break. $<$/s$>$}

\texttt{\textbf{Question:} A cow eats some hay, an apple and a piece of bread. In its tummy}
\texttt{\textbf{Choices:}A: mail box; B: suitcase; C: front door; D: bowl; E: post office}

\texttt{\textbf{Answer:} B (suitcase)}  

\texttt{\textbf{Rationale:}The answer is B because the cow's stomach contains digestive enzymes that break down the consumed food into smaller, soluble molecules through the process of dissolution.$<$/s$>$}
\end{flushleft}

\subsection*{QuaRel}
\begin{flushleft}
\small
\texttt{$<$s$>$$[$INST$]$ Given the following two examples of question-answer-rationale triplets, provide a rationale for the third example for why the selected choice answers the question. $[$\textbackslash INST$]$}  

\texttt{\textbf{Question:} Dan drives a car into the garage from the gravel parking lot. The car moves more smoothly into the garage than the parking lot. This is because there is a bumpier surface in the (A) garage floor (B) gravel parking lot.}

\texttt{\textbf{Answer:} B (gravel parking lot)}  

\texttt{\textbf{Rationale:} The answer is B because the question states that the car moves more smoothly into the garage than the parking lot, indicating that the gravel parking lot has a bumpier surface compared to the garage floor. $<$/s$>$}

\texttt{\textbf{Question:} he baseball team coach was considering both Ted and Roy for the right field position. He needed someone who could propel the ball all the way to the basemen and he knew Ted was more physically fit and possessed greater physical strength than Roy. Who could likely throw the ball a further distance? (A) Roy (B) Ted}

\texttt{\textbf{Answer:} B (Ted)}  

\texttt{\textbf{Rationale:}The answer is B because the question indicates that Ted is more physically fit and possesses greater physical strength than Roy, suggesting that Ted is more likely to throw the ball a further distance.$<$/s$>$}
\end{flushleft}

\section{Models and Reproducibility}
We used the Huggingface library (v4.26.1; \citealt{wolf-etal-2020-transformers}) and publicly available checkpoints for both student\footnote{\url{https://huggingface.co/openai-community/gpt2-xl}; \url{https://huggingface.co/microsoft/phi-1_5};\url{https://huggingface.co/google/gemma-2b}} and teacher\footnote{\url{https://huggingface.co/mistralai/Mistral-7B-Instruct-v0.2}} models. GPT-2 and Phi-1.5 were fine-tuned on a single A100 instance while Gemma-2B was fine-tuned on 2 A100 instances. To monitor the training process, we evaluated model checkpoints every 500 steps. Early stopping was employed with a patience parameter of 10, meaning that training was halted if there was no improvement in the evaluation-set accuracy for 10 consecutive evaluations. The improvement threshold was set to 0.02, ensuring that only significant improvements were considered to continue training. This strategy helped to prevent overfitting and reduced unnecessary computational overhead. Upon publication, we release all code (included elicited rationales for all datasets considered) necessary for reproducing our experiments. 

\section{RQ1 example heatmaps}
\label{appx:lensviz}
Now we visualize the predictions of individual layers of GPT-2 fine-tuned with no, pre, and post-CoT rationales while processing the input ``Question: an electric car contains a motor that runs on...Choices: A: gas; B: hydrogen; C: ions; D: plutonium'', we are specifically interested in what the layers in the penultimate time-step (w.r.t final target label) \textit{think} the next token should be. 

\begin{figure*}
 \centering
  \includegraphics[angle=270, scale=0.35]{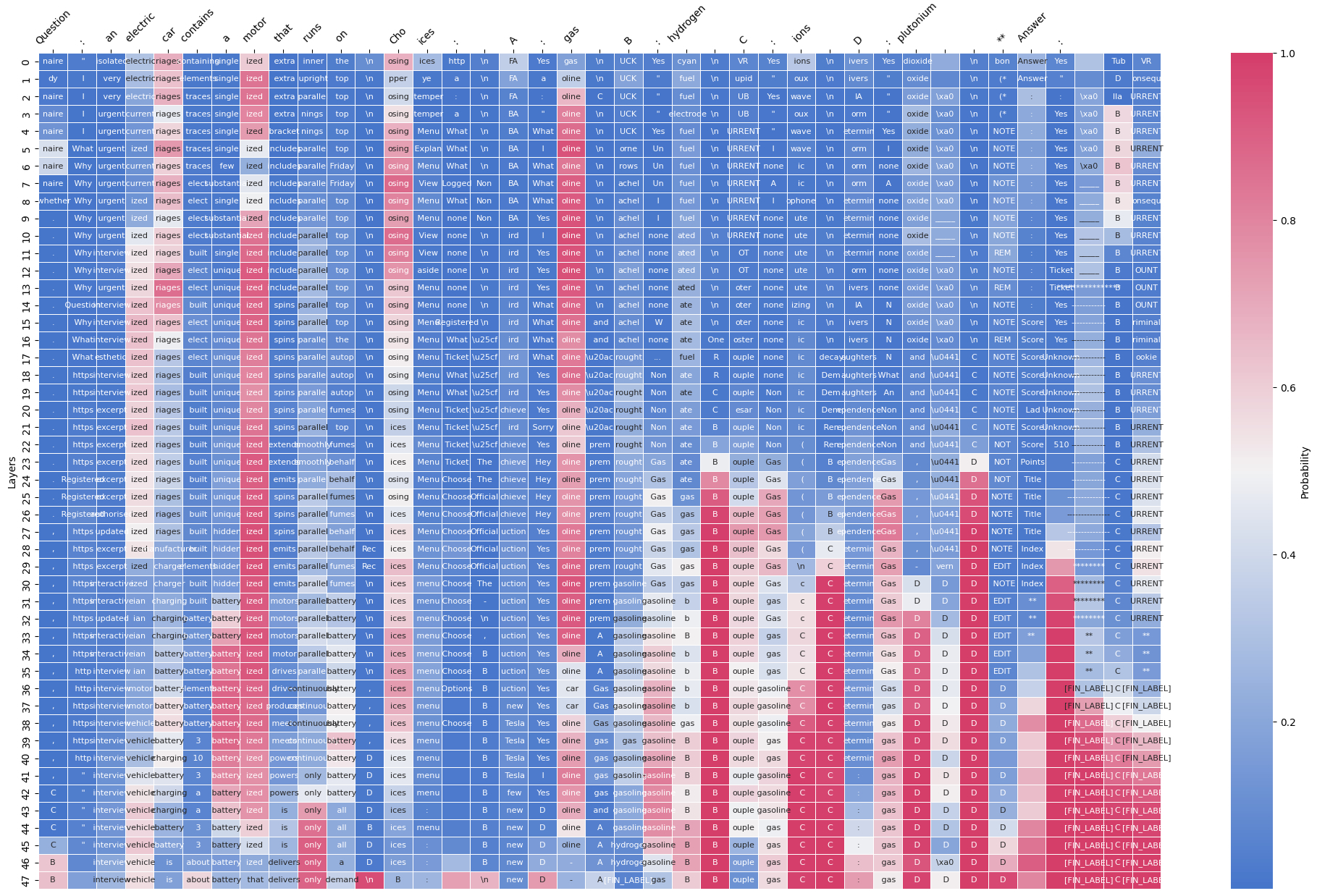}
  \caption{\textbf{No-CoT Baseline}: GPT-2 variant fine-tuned without COT rationales. In this example, the model is confident (p$>$0.6) of the first occurrence of the correct label C at \textbf{layer 40}.}
  \label{fig:apx_lensnocot}
\end{figure*}

\begin{figure*}
\centering
  \includegraphics[angle=270, scale=0.35]{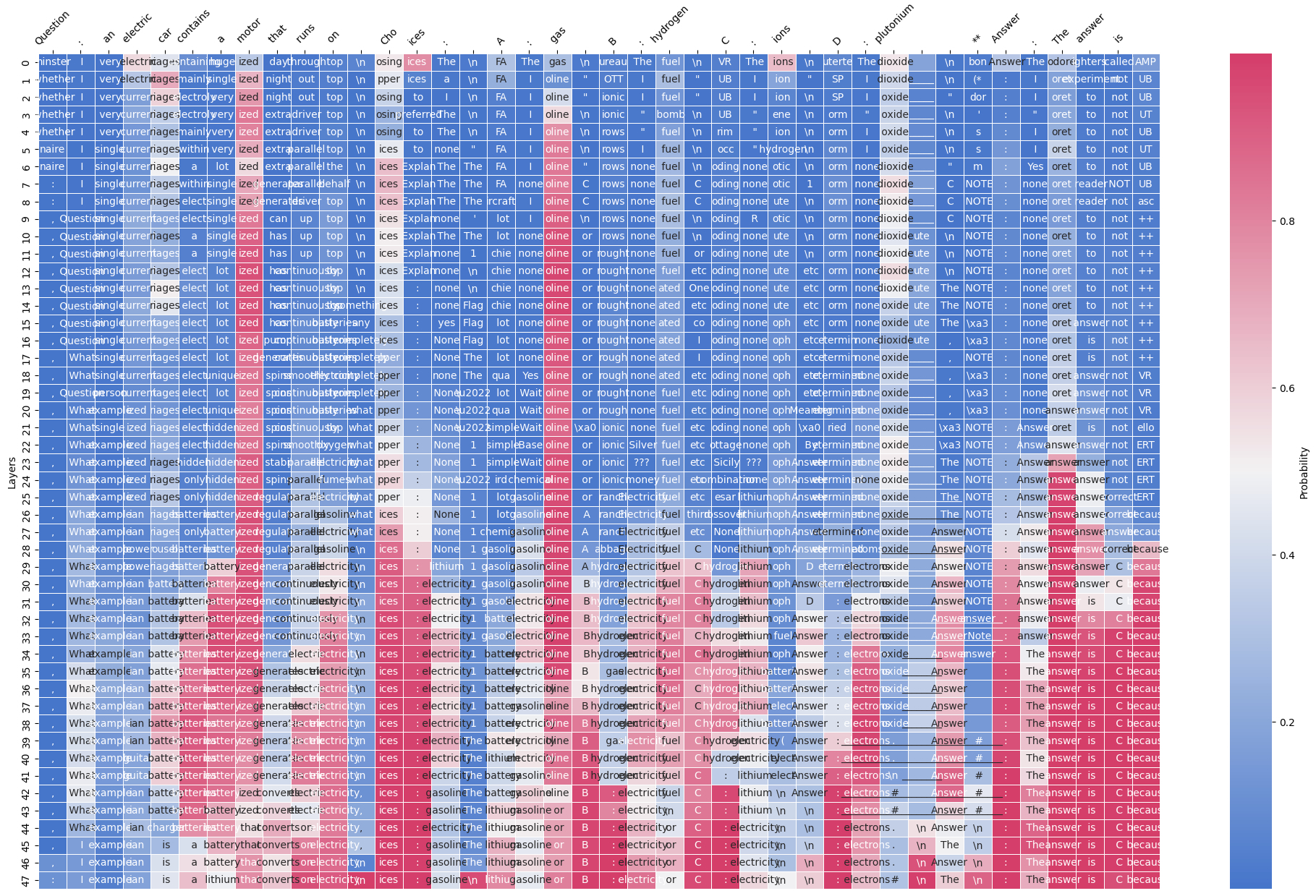}
  \caption{\textbf{Pre-CoT}: GPT-2 variant fine-tuned with CoT rationales \textit{pre-pended} to the target label. In this example, the model is confident (p$>$0.6) of the first occurrence of the correct label C at \textbf{layer 33}.}
  \label{fig:lensprecot}
\end{figure*}

\begin{figure*}
\centering
 \includegraphics[angle=270, scale=0.35]{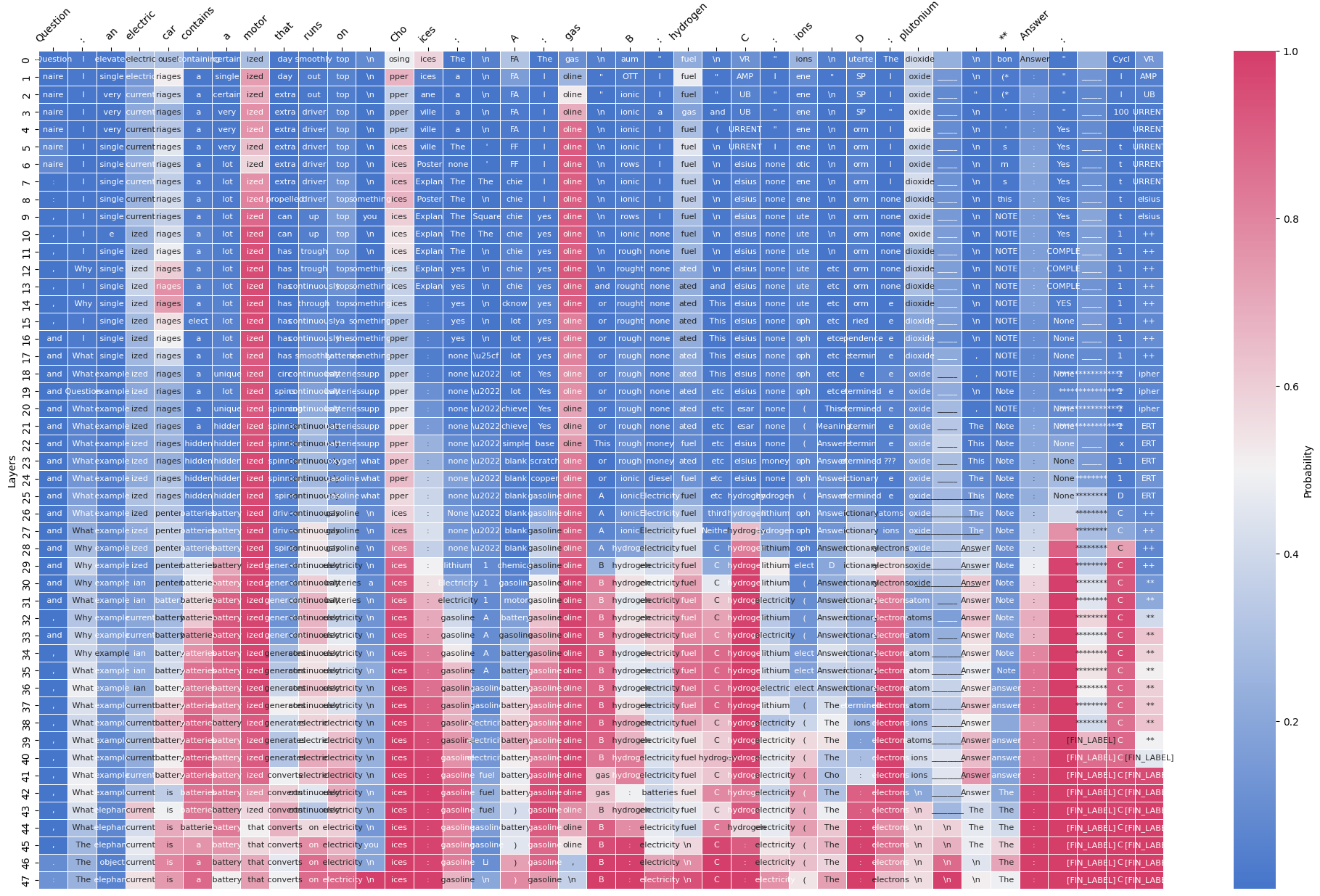}
  \caption{\textbf{Post-CoT}: GPT-2 variant fine-tuned with CoT rationales \textit{appended} to the target label. In this example, the model is confident (p$>$0.6) of the first occurrence of the correct label C at \textbf{layer 27}.}
  \label{fig:lensprecot}
\end{figure*}

\end{document}